\documentclass[runningheads]{llncs}
\usepackage{graphicx}

\usepackage{tikz}
\usepackage{comment}
\usepackage{amsmath,amssymb} 
\usepackage{url}
\usepackage{hyperref}
\usepackage{booktabs}
\usepackage{graphicx}
\usepackage{subfigure}
\usepackage{bm}
\usepackage{color}
\usepackage{amsfonts}
\usepackage{stmaryrd}
\usepackage{stfloats}
\usepackage{xspace}
\usepackage{threeparttable}
\usepackage{multirow}

\newcommand{\ie}{\emph{i.e.,}\xspace}
\newcommand{\eg}{\emph{e.g.,}\xspace}
\newcommand{\etal}{\emph{et al.}\xspace}

\begin{document}
\pagestyle{headings}
\mainmatter

\title{Deep $N$-ary Error Correcting Output Codes} 


\titlerunning{Deep $N$-ary Error Correcting Output Codes}
%
\author{Hao Zhang\inst{1} \and
Joey Tianyi Zhou\inst{1} \and
Tianying Wang\inst{1} \and
Ivor W. Tsang\inst{2} \and
Rick Siow Mong Goh\inst{1}}
\authorrunning{Zhang et al.}
%
\institute{Institute of High Performance Computing, A*STAR, Singapore\\
\and
Australian Artificial Intelligence Institute, UTS, Australia\\
\email{\{zhang\_hao,joey\_zhou,wang\_tianying,gohsm\}@ihpc.a-star.edu.sg}\\
\email{ivor.tsang@uts.edu.au}\\
}
\maketitle

\pagestyle{plain}
\setcounter{footnote}{0}

\begin{abstract}
Ensemble learning consistently improves the performance of multi-class classification through aggregating a series of base classifiers. To this end, data-independent ensemble methods like Error Correcting Output Codes (ECOC) attract increasing attention due to its easiness of implementation and parallelization. Specifically, traditional ECOCs and its general extension $N$-ary ECOC decompose the original multi-class classification problem into a series of independent simpler classification sub-problems. Unfortunately, integrating ECOCs, especially $N$-ary ECOC with deep neural networks, termed as Deep $N$-ary ECOC, is not straightforward and yet fully exploited in the literature, due to the high expense of training base learners. To facilitate the training of $N$-ary ECOC with deep learning base learners, we further propose three different variants of parameter sharing architectures for deep $N$-ary ECOC. To verify the generalization ability of deep $N$-ary ECOC, we conduct experiments by varying the backbone with different deep neural network architectures for both image and text classification tasks. Furthermore, extensive ablation studies on deep $N$-ary ECOC show its superior performance over other deep data-independent ensemble methods.\footnote{Our code is available at \url{https://github.com/IsaacChanghau/DeepNaryECOC}.}

\end{abstract}

\section{Introduction}
Multi-class classification is one of the fundamental problems in machine learning and data mining communities, where one trains a model with labeled data of different classes for classification purposes. The multi-class classification exists diverse real-world applications from computer vision tasks such as object recognition~\cite{felzenszwalb2005pictorial,lowe1999object,riesenhuber1999hierarchical}, face verification~\cite{kittler2003face}, to natural language processing tasks like sentiment classification~\cite{glorot2011domain,pang2002thumbs}.

To handle multi-class classification problems, existing approaches could be mainly divided into two groups. One group focuses on solving the multi-class problems directly by extending its corresponding binary classification algorithm. These approaches include decision tree-based methods~\cite{deng2011fast}, multi-class linear discriminant analysis~\cite{torkkola2001linear}, multi-layer perceptron~\cite{freund1999large}, multi-class support vector machines (SVM)~\cite{chang2011libsvm} and etc. Another research direction focuses on the decomposition of a multi-class problem into multiple binary sub-problems so that one can reuse the well-studied binary classification algorithms for their simplicity and efficiency. Most of these methods can be reinterpreted in the framework of error correcting output codes (ECOC)~\cite{dietterich1991error,dietterich1994solving}. For example, Allwein \etal~\cite{allwein2000reducing} show one-versus-one (OVO), one-versus-all (OVA) could be incorporated into the framework of ECOC where all the classes are reassigned with either binary codes $\{-1,1\}$ or ternary codes $\{-1,0,1\}$ for each base learners ($1/-1$ represents positive/negative class, $0$ represents non-considered class). Zhou \etal~\cite{zhou2019n} further extend traditional ECOCs into $N$-ary ECOC by introducing $N$ meta-classes rather than binary classification for each base learner. The final results are determined by the ensemble of a series of base learners. The biggest advantage of ECOCs methods is their easiness of implementation and parallelization.

Most traditional ECOCs methods are based on the pre-defined hand-craft features and focus on how to ensemble the results of base learners on these features. Recently, deep learning methods significantly advance the multi-class classification performance through learning features in an end-to-end fashion. For example, a single AlexNet~\cite{alex2012imagenet} outperforms the second place at the ImageNet Large Scale Visual Recognition Challenge (ILSVRC) by more than $10\%$. To further improve performance, Goodfellow \etal~\cite{goodfellow2016deep} demonstrate that a simple ensemble of seven AlexNet models with different random initiations could significantly reduce an error rate from $18.2\%$ to $15.3\%$. In most high profile competitions, e.g. ImageNet\footnote{ImageNet: \url{http://www.image-net.org/}}~\cite{deng2009imagenet} or Kaggle\footnote{Kaggle: \url{https://www.kaggle.com/}}, ensembles techniques often appear in the winner solution. Traditional ensemble methods usually assume that the base learners for binary classification are inexpensive to train, such as SVMs and decision trees. Unfortunately, this assumption appears to be invalid with deep learning algorithms. For example, AlexNet consisting of more than 60 millions of parameters~\cite{alex2012imagenet} takes between five and six days to train on two GTX 580 3GB GPUs. Therefore, the expensive learning procedure hinders the use of the ensemble of deep neural networks on a large scale.

In this paper, we focus on addressing the ensemble of deep neural networks in the
framework of ECOC. The biggest reason to choose ECOC rather than other ensemble
techniques such as Boosting [18] is that ECOC is easy to parallel due to the independence of base learners. In contrast, boosting trains a number of models sequentially and continuously compensates the mistakes made by the earlier models, which results in that each base models in boosting are highly dependent on each other. At this point, ECOC exhibits a large advantage in large-scale real-world applications since all the base learners could be trained independently and simultaneously.

Specifically, we choose $N$-ary ECOC, an extension of ECOC, which shows significant
improvement over OVA, OVO, and traditional ECOCs~\cite{zhou2019n}. Many existing works did not investigate the influence of deep learning on ECOCs or $N$-ary ECOC. In this paper, we make a marriage between $N$-ary ECOC to investigate such an influence. In the sequence, we term this problem as Deep $N$-ary ECOC. The main contributions of this paper are as follows:
\begin{itemize}
    \item We investigate a new problem named \textit{Deep $N$-ary ECOC} where we mainly discuss how to effectively and efficiently leverage advantages of deep learning models in the framework of ECOC.
    \item To facilitate the training procedure, we further propose three different parameter sharing strategies for Deep $N$-ary ECOC framework, \ie full parameters share, partial parameters share, and no parameter share. Specifically, the full share model shares all the feature learning parameters except for top classifier; the partial share model shares part of feature learning parameters; the no parameter share means all the base learners are learned from scratch.
    \item We explore the influence of two crucial hyper-parameters of $N$-ary ECOC, \ie $N_L$ and $N$, with deep neural networks for improving the accuracy. We also give specific suggestions for choosing those two hyper-parameters.
    \item We conduct extensive experiments and compare with several ensemble strategies, \ie an ensemble of random initialization (ERI), ECOC and $N$-ary ECOC, on both image and text classification tasks to analyze the advantages and disadvantages of each ensemble strategy.
\end{itemize}

The rest of this paper is organized as follows. Section~\ref{sec:related_work} reviews related work. Section~\ref{sec:deep_nary_ecoc} presents Deep $N$-ary ECOC. Finally, Section~\ref{sec:experiment} discusses our empirical studies and Section~\ref{sec:conclude} concludes this work.

\section{Related Work}\label{sec:related_work}
Our proposed deep $N$-ary ECOC is highly related to the following topics, including
ECOCs, ensemble learning, and deep neural networks.

\subsection{ECOCs}
Many ECOC approaches~\cite{allwein2000reducing,bagheri2013subspace,escalera2008decoding,pujol2006discriminant} have been proposed to design a good coding matrix in recent years. Most of them are fallen into the following two categories. The first one is data-independent coding, such as OVO, OVA, and ECOCs~\cite{escalera2008decoding}. Their coding matrix design is not optimized for the training dataset nor the instance labels such that all the base learners could be independently learned. For example, the sparse ECOC coding approach aims to construct the ECOC matrix $M\in\{-1, 0, 1\}^{N_C\times N_L}$, where $N_C$ is the number of classes, $N_L$ is the code length, and its elements are randomly chosen as either $-1$, $1$, or $0$~\cite{escalera2008decoding}. In ECOCs, the classes corresponding to $1$, $-1$ are considered as positive and negative classes, respectively, and $0$ are not considered in the learning process. More recently, Zhou \etal~\cite{zhou2019n} extend the existing ECOCs into $N$-ary ECOC to enable the construction of $N$ meta-classes. Both theoretical and empirical findings validate the superiority of $N$-ary ECOC over traditional ECOCs.

Another direction is data-dependent ECOCs where the data are considered in the learning coding matrix, such as discriminant ECOC (D-ECOC)~\cite{pujol2006discriminant}, ECOC-ONE~\cite{radeva2006ecoc}, subspace ECOC~\cite{bagheri2013subspace}, Adaptive ECOC~\cite{zhong2013adaptive}, etc. In this way, different base learners interact with each other during training phrases, which is also similar to Boosting~\cite{schapire1990strength} methods, such as AdaBoost~\cite{freund1997decision}. In Boosting methods, a series of models are sequentially trained with latter models correcting mistakes committed in previous models. Compared to data-independent ECOCs, these methods require sophisticated algorithm design and are difficult to be paralleled.

To our best knowledge, there is little research to investigate the combination of ECOCs and deep learning. In this paper, we take a step further to analyze the performance of combining our previous work $N$-ary ECOC with deep learning.

\subsection{Deep Ensemble Learning}
A lot of studies show that deep neural network models are nonlinear and have a high variance, which can be frustrating when preparing a final model for making predictions~\cite{goodfellow2016deep}. Deep ensemble learning appears to one of the solutions that combine the predictions from multiple neural network models to reduce the variance of predictions and reduce generalization error. Recently, there are some studies to integrate base learners of deep neural networks with ensemble learning in three major ways. The first one is ensemble training data including re-sampling~\cite{efron1982jackknife}, bootstrap aggregation~\cite{breiman1996bagging}, where the choice of data is varied for training different base models in the ensemble. The second one is to ensemble models where different base models are used in the ensemble, including different random initialization, a random selection of mini-batches, differences in hyper-parameters, etc~\cite{goodfellow2016deep}. The third way is varying combinations where one vary the choice of combining outcomes from ensemble members. The most famous method is a model averaging ensemble and weighted average ensemble. Different from the aforementioned deep ensemble learning methods, deep $N$-ary ECOC serves a complementary piece for existing methods.

\subsection{Deep Neural Networks}
In recent years, a lot of different deep neural networks are proposed for different applications. For computer vision tasks, the most dominating model comes from Convolutional Neural Networks (CNNs), and its follow-up works such as AlexNet~\cite{alex2012imagenet}, VGGs~\cite{simonyan2014very}, ResNet~\cite{he2016deep} and DenseNet~\cite{huang2017densely}. For natural language processing tasks, most popular networks belong to Recurrent Neural Network (RNNs), or its many variants such as Long Short-Term Memory (LSTM)~\cite{hochreiter1997long}, Gated Recurrent Unit (GRU)~\cite{cho2014learning} and etc. In the experiment, we validate deep $N$-ary ECOC in both CNNs and LSTMs architecture for vision and text datasets, respectively.

\section{Deep $N$-ary ECOC}\label{sec:deep_nary_ecoc}
In this section, we first introduce the concept of $N$-ary ECOCs. To facilitate the training procedure, we further propose three different parameter sharing architectures, namely full, partial and no sharing.

\begin{figure}[t]
    \centering
	\subfigure[\small ECOC]
	{\label{ecoc_example}	\includegraphics[width=0.45\textwidth]{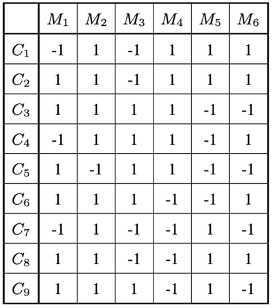}}
	\subfigure[\small $N$-ary ECOC with $N=4$]
	{\label{nary_ecoc_example}	\includegraphics[width=0.45\textwidth]{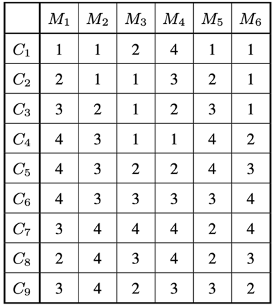}}
	\caption{\small Example of ECOC and $N$-ary coding matrix.}
	\label{fig:ecoc_nary_ecoc_example}
\end{figure}

\subsection{$N$-ary Ensemble for Multi-class Classification}
Error correcting output codes (ECOC) constructs an ensemble of binary base classifiers by randomly and independently assigning positive/negative pseudo labels (\ie $1/-1$ in the coding matrix) for each base task. The results of all the base learners are combined to make a prediction. ECOC consists of two main steps: 1) encoding 2) decoding. In encoding, we create an encoding matrix to encode each class into a unique code that is as different as possible from the codes of the remaining classes. One example of the encoding matrix is illustrated in Fig.~\ref{ecoc_example}. A row of the coding matrix represents the code of each class, while a column of the coding matrix represents the binary classes to be considered when learning a base classifier. In decoding, ECOC first computes the prediction vector that is the concatenation of results of all the base tasks. The final label is determined by assigning the label with the “closest” label vector in encoding matrix $M\in\{1,2,\dots,N\}^{N_C\times N_L}$, where $N_C$ denotes the number of classes and $N_L$ denotes the number of base learners. As proved in many research works~\cite{dietterich1994solving,zhou2019n}, the capability of error correction relies on the minimum distance, $\Delta_{\min}(M)$, between any distinct pair of rows in the coding matrix $M$. In this way, the trained base classifiers could be sufficiently differentiated from each other.

To achieve this goal, ECOC is extended to a new framework named $N$-ary ECOC~\cite{zhou2019n}, where the original classes are decomposed into $N$ meta-class ($3\leq N \leq N_C$). Fig.~\ref{nary_ecoc_example} shows an example of $N$-ary ECOC encoding matrix. Zhou \etal~\cite{zhou2019n} both empirically and theoretically showed that $N$-ary ECOC is able to achieve larger row separation and lower column correlation. It is interesting to note that $N$-ary ECOC is a more general framework for ECOC since traditional coding schemes could be treated as special cases of $N$-ary ECOC. For
example, when $N=2$, $N$-ary ECOC corresponds to the binary coding scheme; when $N=3$, $N$-ary ECOC corresponds to the ternary coding scheme. Furthermore, recent works~\cite{goodfellow2016deep} showed that an Ensemble of models with different Random Initialization (ERI) is able to improve multi-class classification performance. This deep ensemble learning strategy could be also viewed as a special case in the framework of $N$-ary ECOC if we keep the original label assignment, namely $N=N_C$.

On the other side, most existing work on ECOCs including our previous work on $N$-ary ECOC is constrained to classifier training with pre-defined features. With a significant advance of deep learning, the performance of various machine learning tasks has been improved. There is few works to discuss how to extend ECOC in the scenario of deep learning. In this work, we specifically study this open problem in the framework of $N$-ary ECOC, termed \textit{Deep $N$-ary ECOC}, and propose several approaches to address it. In this paper, we mainly investigate the following three questions:
\begin{enumerate}
    \item Do we necessarily independently train all the deep base learners from scratch for all the situation?
    \item Whether the $N$-ary ECOC framework still retains the advantages over other data-independent ensemble approaches with deep neural network?
    \item Any new suggestion on the choice of the meta-class number $N$ and base learners number $N_L$?
\end{enumerate}

For the first question, we are going to propose three different parameter sharing architectures, which is described in more details next section. For the remaining two questions, we delay the investigation in the experiment section.

\subsection{Efficient Implementation for Deep $N$-ary ECOCs}
\begin{figure}[t]
    \centering
	\includegraphics[width=0.9\textwidth]{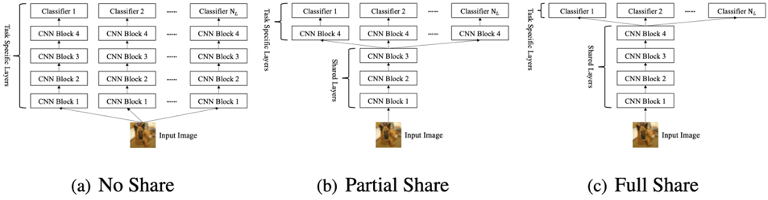}
	\caption{\small An example of three different parameters sharing strategies.}
	\label{fig:param_sharing_architecture}
\end{figure}

Different from the traditional ECOC with pre-defined features, deep ECOCs require to consider deep feature learning as well as the classifier construction during training. This increases the difficulty of deploying ECOCs in real-world scenarios, since even training a single deep neural network is also expensive. Fortunately, thanks to the nature of ECOCs, all the base deep neural networks could be trained simultaneously. Furthermore, in this paper, we investigate a more efficient realization and propose three different parameter sharing strategies, namely, no share, partial share, and full share, which is depicted in Fig.~\ref{fig:param_sharing_architecture}.

Typically, we take the model of the CIFAR dataset, termed CIFAR-CNNs (explain in detail later), as an example to illustrate the three strategies. For the no parameter sharing strategy, as shown in Fig.~\ref{fig:param_sharing_architecture}(a), we trained $N_L$ base learners independently, which means that the feature encode layers of each base learner are trained by the inputs directly and do not interact with other learners. The partial parameter sharing strategy contains shared and task-specific layers, as in Fig.~\ref{fig:param_sharing_architecture}(b), the first three feature encoder layers are shared by all the base learners while the top encoder layer is task-specific, which is only optimized by the corresponding meta-class objectives. The full parameter sharing strategy is simply set all the feature encode layers to be shared by all the base learners except the top classifiers (see Fig.~\ref{fig:param_sharing_architecture}(c)). The top layer classifiers of all the sharing strategies are trained independently with its meta-class objectives. Note that, all the base learners of no parameters sharing strategy are trained from scratch while the shared layers of partial and full parameters sharing strategies are initialized by a pre-trained single model and fine-tuned through training to accelerate the model convergence rate. Obviously, the no parameter sharing strategy contains most parameters ($N_n$), then the partial sharing strategy ($N_p$) and the full sharing strategy ($N_f$) is least, say, $N_n>N_p>N_f$.

\section{Experiments}\label{sec:experiment}

\subsection{Datasets}
We conduct the experiments on 4 image datasets and 2 text datasets. The image datasets contain MNIST~\cite{lecun1998gradient}, CIFAR-10~\cite{krizhevsky2009learning}, CIFAR-100~\cite{krizhevsky2009learning}, and FLOWER-102~\cite{nilsback2008automated}, which are widely used image classification datasets in the computer vision community. The text datasets are Text REtrieval Conference (TREC)~\cite{li2002learning} dataset and Stanford Sentiment Treebank (SST)~\cite{socher2013recursive} dataset. The TREC is the question and answering dataset which involves classifying question sentences into 6 question types, say, whether the question is about person, location, numeric information and etc. The SST is the sentiment analysis sentence data with $5$ classes that range from $0$ (most negative) to $5$ (most positive). The statistics of these datasets are described in Table~\ref{tab:stat_datasets}. Note that we do not utilize the $K$-fold cross-validation method, but simply use the split of train/validation/test sets. If the datasets do not contain development part, we randomly split $10\%$ training samples as the development dataset.

\begin{table}[t]
    \scriptsize
    \caption{\small Statistics of Image and Text Datasets.}
	\label{tab:stat_datasets}
	\centering
	\begin{tabular}{l c c c c c}
		\toprule
		\multicolumn{6}{c}{Image Dataset} \\
		\midrule
        Dataset & Image Size & \# Train Samples & \# Dev Samples & \# Test Samples & \# Classes ($N_C$) \\
        \midrule
        MNIST & $28\times 28$ & $60,000$ & N/A & $10,000$ & $10$ \\
        CIFAR-10 & $32\times 32$ & $50,000$ & N/A & $10,000$ & $10$ \\
        CIFAR-100 & $32\times 32$ & $50,000$ & N/A & $10,000$ & $100$ \\
        FLOWER-102 & $256\times 256$ & $6,552$ & $818$ & $819$ & $102$ \\
        \midrule
        \multicolumn{6}{c}{Text Dataset} \\
        \midrule
        Dataset & Avg. Sent. Len. & \# Train & \# Dev & \# Test & \# Classes ($N_C$) \\
        \midrule
        TREC & $10$ & $5,500$ & N/A & $500$ & $6$ \\
        SST & $18$ & $11,855$ & N/A & $2,210$ & $5$ \\
        \bottomrule
	\end{tabular}
\end{table}

\subsection{Experimental Setup}
\subsubsection{Deep Neural Networks.} 
\begin{figure}[t]
    \centering
	\includegraphics[width=0.9\textwidth]{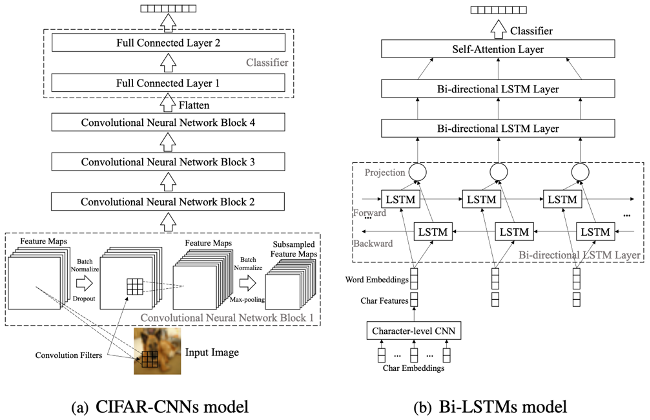}
	\caption{\small The general architecture of \textit{CIFAR-CNNs} and \textit{Bi-LSTMs} models.}
	\label{fig:general_architecture}
\end{figure}

We employ different neural network-based models for different datasets. Specifically, we use LeNet~\cite{lecun1998gradient} for the MNIST dataset and the FLOWER-102 dataset is trained by AlexNet~\cite{alex2012imagenet}. Note that, due to the difficulty for the AlexNet model to learn consequential and representative features from the small training dataset of FLOWER-102 directly, the AlexNet is not trained from scratch but obtained by fine-tuning the pre-trained AlexNet model\footnote{Pre-trained AlexNet: \url{http://www.cs.toronto.edu/~guerzhoy/tf_alexnet/}}, which is trained on ILSVRC dataset. For CIFAR-10/100 datasets, we build a model with eight convolutional layers and two full-connected layers, named as \textit{CIFAR-CNNs}, as shown in Fig.~\ref{fig:general_architecture}(a), where the eight convolutional layers are divided into four groups, they share the same structure with the different numbers of filters and kernel widths. The architecture of each group is structured as follows: one convolutional layer following the batch normalization~\cite{ioffe2015batch} and dropout~\cite{srivastava2014dropout} layer, another convolutional layer with batch normalization and max-pooling is applied. And the ELU~\cite{clevert2016fast} activation function is used for each convolutional layer.

To train the TREC and SST text datasets, we construct a three-layer bidirectional LSTM model with character-level CNN~\cite{kim2016character} and self-attention~\cite{bahdanau2015neural} mechanism, termed \textit{Bi-LSTMs}, as shown in Fig.~\ref{fig:general_architecture}(b), where the character-level CNN learned the character features to represent a word from the character sequences of such word, which can help to enrich the meaning of word features, especially for rare and out-of-vocabulary words, and boost the performance by capturing morphological and semantic information, and the self-attention mechanism encodes the learned contextual affluent word-level feature sequence of bidirectional LSTM into a single vector by considering the importance of each word feature.

\subsubsection{Parameters Setup.} For LeNet of MNIST dataset, we follow the same settings as LeCun \etal~\cite{lecun1998gradient} and RMSProp~\cite{tieleman2012lecture} is chosen as the parameters optimization method with a learning rate of $0.001$ and decay rate of $0.9$, we also introduce Dropout~\cite{srivastava2014dropout} strategy with a drop rate of $0.5$ at each convolutional layer and the first full-connected layer to prevent over-fitting. While the AlexNet for FLOWER-102 dataset, we utilize exactly the same structure, parameter setting, and optimization method as Alex \etal~\cite{alex2012imagenet}. For \textit{CIFAR-CNNs} model of CIFAR-10/100 datasets, we set the number of filters for each convolutional block as $32$, $64$, $128$, and $256$, respectively, the kernel sizes of $(3,3)$ and pool size of $(2,2)$ for all the blocks. The hidden size of the first fully-connected layer is $512$, while the second depends on the class size. To avoid over-fitting, we apply $l_2$ regularization with weight decay rate of $0.0005$ for all the weight parameters and Dropout strategy, the drop rate is $0.3$, $0.4$, $0.4$, $0.4$ for each convolutional block respectively, and $0.5$ for the first fully-connected layer. Parameters optimization is performed by Adam optimizer~\cite{kingma2015adam} with gradient clipping of $5.0$ and learning rate decay strategy. We set the initial learning rate of $\beta_0=0.002$ and fixed it for the first $5000$ training iterations, then the learning rate $\beta_t$ is updated by $\beta_t=\beta_0 / \big(1 + \rho \times \frac{t-5000}{T}\big)$, where $T$ is the decay step of $500$ and $\rho$ is the decay rate of $0.05$. Meanwhile, in order to improve the performance, the data augmentation is also utilized.

For the \textit{Bi-LSTMs} model of TREC and SST text datasets, we use the $300$-dimensional publicly available pre-trained word embeddings as the word-level feature representation, which is trained by fastText\footnote{fastText: \url{https://github.com/facebookresearch/fastText}} package on \textit{Common Crawl} and \textit{Wikipedia}~\cite{bojanowski2017enriching,grave2018learning}, and the $50$-dimensional randomly initialized task-specific character embeddings. The word embeddings are fixed and character embeddings are learned during training. We use three different convolutional layers with widths $2$, $3$, $4$, respectively, for character-level CNN encoder and set the filter number of each layer as $20$, the learned character features of each layer are concatenated and then optimized by a two-layer highway network~\cite{srivastava2015highway} before concatenating with the corresponding word embeddings. The dimension of hidden states of LSTM layers are set as $200$. Parameters optimization is performed by Adam optimizer~\cite{kingma2015adam} with gradient clipping of $5.0$ and learning rate decay strategy. We set the initial learning rate of $\beta_0=0.001$, at each epoch $t$, learning rate $\beta_t$ is updated by $\beta_t=\beta_0/(1+\rho\times t)$, where $\rho$ is the decay rate with $0.05$. To reduce overfitting, we also apply Dropout~\cite{srivastava2014dropout} at the embedding layer and the output of each LSTM layer with the drop rate of $0.2$ and $0.3$, respectively.

\subsubsection{$N$-ary ECOC Coding Matrix Setup.} For $N$-ary ECOCs, including ECOCs (which is a special issue of $N$-ary ECOC when $N=2$), we train the $N_L$ base learners based on the coding matrix and use the predicted code sequence of each class and generated coding matrix to make a prediction based on distance measurement. Zhou \etal~\cite{zhou2019n} introduced several coding matrix construction methods and distance measurements designed for general or task-specific applications. For simplicity, we utilize the random dense encoding method to randomly split the original classes $N_C$ into $N$ subsets and make sure that the number of classes in each subset should be approximately balanced, simultaneously. For the decoding method, we adopt the minimum Hamming distance due to its simplicity and effectiveness. In our experiments, we experiment on the different numbers of meta-class $N$ and number of base learners $N_L$ for different datasets, as described in Table~\ref{tab:summ_n_n_l}. Note that we do not experiment on all the possible meta-classes for each dataset, because of the limitations of computing resources and we only trained $60$ base learners for MNIST, FLOWER-102, TREC, and SST datasets and $100$ base learners for CIFAR-10/100 datasets, respectively. Specifically, in order to evaluate the effects of number of base learners on the ensemble learning performance, we trained another $300$ classifiers for FLOWER-102 and TREC datasets, respectively.

\begin{table}[t]
    \scriptsize
    \caption{\small Summarization of tested $N$ and $N_L$ for experiments.}
	\label{tab:summ_n_n_l}
	\centering
	\begin{threeparttable}
	\begin{tabular}{l c c c}
		\toprule
        Dataset & \# Classes ($N_C$) & Tested \# Meta-Class ($N$) & Tested \# Base Learners* ($N_L$) \\
        \midrule
        MNIST & $10$ & $2,4,5,8,10$ & $60$ \\
        CIFAR-10 & $10$ & $2,4,5,8,10$ & $100$ \\
        CIFAR-100 & $100$ & $2,5,10,30,50,75,95,100$ & $100$ \\
        FLOWER-102 & $102$ & $2,3,5,10,20,40,60,80,90,95,102$ & $60$ \\
        \midrule
        TREC & $6$ & $2,3,4,5,6$ & $60$ \\
        SST & $5$ & $2,3,4,5$ & $60$ \\
        \bottomrule
	\end{tabular}
	\begin{tablenotes}
        \small
        \item[] \scriptsize{*It indicates the maximal number of classifiers is used for training.}
      \end{tablenotes}
     \end{threeparttable}
\end{table}

\subsection{Experimental Results}

\subsubsection{Comparison with Different Ensemble Methods.}
In this section, we compare the performance of different ensemble methods on the aforementioned image and text datasets. In the experiment, we trained a single model and ensemble models of three coding schemes for each dataset, \ie Ensemble with Random Initializations (ERI), ECOC, $N$-ary ECOC, then report their (ensemble) accuracy with standard deviations. Note that we only report the highest score under a specific meta-class $N$ for $N$-ary ECOC. For the MNIST, FLOWER-102, TREC, and SST datasets, we use $60$ base learners for each scheme, while $100$ base learners for CIFAR-10/100 datasets. The results are summarized in Table~\ref{tab:main_results}. Generally, we observe that most ensemble models show relatively significant improvements, compared with the single model, on the given datasets with different deep neural networks.

We observe two interesting results in Table~\ref{tab:main_results}. First, comparing the single model with $N$-ary ECOC, we find that the improvement ratio of $N$-ary ECOC is inverse relation with single model performance, \ie the improvement of $N$-ary ECOC scheme is more prominent if the performance of the single model is lower. For example, it is obvious that the baseline accuracies are higher on MNIST, CIFAR-10, TREC and FLOWER-102 ($>80\%$) than on CIFAR-100 and SST (almost $<60\%$), then the improvement ratios are $0.59\%$, $5.54\%$, $5.64\%$ and $5.80\%$ from the single model to $N$-ary ECOC on MNIST, CIFAR-10, TREC, and FLOWER-102 datasets, respectively, while the improvement ratios of CIFAR-100 dataset are $13.72\%$ and $15.15\%$ on SST dataset.

\begin{table}[t]
    \scriptsize
    \caption{\small Ensemble accuracies with their standard deviations.}
	\label{tab:main_results}
	\setlength{\tabcolsep}{5.5 pt}
	\centering
	\begin{threeparttable}
	\begin{tabular}{l l c c c c}
		\toprule
        \multirow{2}{*}{Dataset} & \multirow{2}{*}{Method} & \multirow{2}{*}{Single Model} & \multicolumn{3}{c}{Ensemble Model*} \\
        & & & ERI & ECOC & $N$-ary ECOC \\
        \midrule
        MNIST & LeNet~\cite{lecun1998gradient} & 98.98$\pm$0.07\% & 99.11$\pm$0.11\% & 99.23$\pm$0.08\% & \textbf{99.57}$\pm$0.09\% \\
        CIFAR-10 & CIFAR-CNNs & 87.12$\pm$0.43\% & 90.54$\pm$0.31\% & 89.37$\pm$0.54\% & \textbf{91.95}$\pm$0.24\% \\
        CIFAR-100 & CIFAR-CNNs & 61.50$\pm$0.57\% & 69.57$\pm$0.29\% & 34.26$\pm$2.42\% & \textbf{69.94}$\pm$0.32\% \\
        FLOWER-102 & AlexNet~\cite{alex2012imagenet} & 83.12$\pm$0.29\% & 86.32$\pm$0.60\% & 77.05$\pm$0.73\% & \textbf{87.94}$\pm$0.28\% \\
        \midrule
        TREC & Bi-LSTMs & 90.50$\pm$0.12\% & 94.80$\pm$0.09\% & 95.80$\pm$0.08\% & \textbf{95.60}$\pm$0.10\% \\
        SST & Bi-LSTMs & 44.17$\pm$0.92\% & 48.69$\pm$0.18\% & 48.91$\pm$0.26\% & \textbf{50.86}$\pm$0.13\% \\
        \bottomrule
	\end{tabular}
	\begin{tablenotes}
        \small
        \item[] \scriptsize{*Here $N_L$ are $60$, $100$, $100$, $60$, $60$ and $60$, respectively, for the ensemble models from top to bottom row. While $N$ are $3$, $4$, $95$, $95$, $3$, $4$, respectively, for the $N$-ary ECOC.}
      \end{tablenotes}
     \end{threeparttable}
\end{table}

Second, the $N$-ary ECOC scheme outperforms ECOC and ERI ensemble methods on most image and text datasets, except for the TREC text dataset. Specifically, $N$-ary ECOC always performs better than ERI. This is due to that $N$-ary ECOC varies the predicted classes for each base learner and makes them more diverse than ERI, where the diverse forecast errors made by base learners of $N$-ary ECOC are more beneficial to the ensemble learning in comparison to the similar base learner errors of ERI. Meanwhile, compared with ECOC, $N$-ary ECOC also shows its superiority in most cases, especially when the number of classes is large (\ie $N_C\geq 100$ in our experiments). It is primarily due to, as mentioned by Zhou \etal~\cite{zhou2019n}, the better quality of the coding matrix and the higher discriminative ability (in terms of how many meta-classes a base learner tries to discriminative) of $N$-ary ECOC than ECOC.

In fact, we find the contribution of class merge degree to the ensemble accuracy of $N$-ary ECOC replies on the dataset, say, the datasets with a different number of classes require different class merge degree strategy, as discuss in Section~\ref{sssec:meta_class}. Note the class merge degree, which is measured by $\frac{N_C-N}{N_C}$ , is the ratio of class numbers reduced when the classes are merged into meta-classes.

\subsubsection{Evaluation on the Effect of Meta-class Number $N$.}\label{sssec:meta_class}
\begin{figure}[t]
    \centering
	\includegraphics[width=0.9\textwidth]{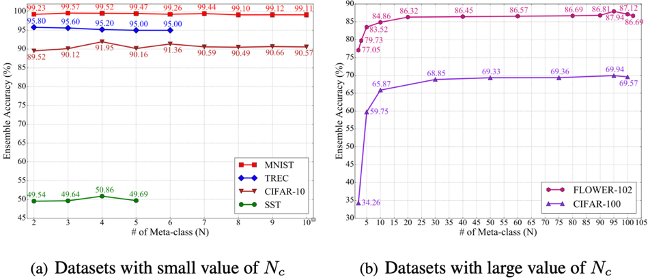}
	\caption{\small Ensemble accuracies with respect to $N$, where the first point of each line represents ECOC ($N=2$), the last represents ERI ($N=N_C$) and the rest is $N$-ary ECOC with various $N$.}
	\label{fig:ensemble_acc_wrt_n}
\end{figure}

In this section, we investigate the influence of meta-class number $N$, which is one of the crucial hyper-parameters of $N$-ary ECOC. For the datasets with a small value of $N_C$, we experiment on all the possible meta-class numbers, \ie from $2$ to $N_C$ ($N=2$ denotes ECOC and $N=N_C$ denotes ERI), while for the datasets with a large value of $N_C$, we select several representative meta-class numbers for the experiment. The ensemble accuracies with respect to $N$ are depicted in Fig.~\ref{fig:ensemble_acc_wrt_n}.

From Fig.~\ref{fig:ensemble_acc_wrt_n}(a), we observe that the performances of ensemble models with different $N$ are relatively stable, the highest ensemble accuracies of MNIST, CIFAR-10, and SST achieve when $N=3$, $N=4$, and $N=4$ respectively, and the best performance of TREC is obtained at $N=3$ if we do not consider the ECOC. After that, the performance of each dataset is gradually decreased with small fluctuations with an increasing number of meta-class $N$. It is interesting to see that $N$-ary ECOC for datasets with a small value of $N_C$ always tends to arrive the best performance with small value of $N$, \ie large class merge degree. Specifically, the class merge degree for MNIST, CIFAR-10, TREC and SST are $0.7$, $0.6$, $0.5$ and $0.2$ respectively.

However, as shown in Fig.~\ref{fig:ensemble_acc_wrt_n}(b), the performance of ensemble models with different $N$ fluctuates significantly on the datasets with a large value of $N_C$. For the ECOC scheme, it only achieves $34.26\%$ ensemble accuracy on the CIFAR-100 dataset and $77.05\%$ on the FLOWER-102 dataset. For the $N$-ary ECOC with $N=3$, it obtains $83.52\%$ and $59.75\%$ on FLOWER-102 and CIFAR-100 datasets, respectively. Then the ensemble accuracy improves gradually with the increase of meta-class $N$ and reaches the summit with accuracies of $87.94\%$ and $69.94\%$, when $N=95$, for FLOWER-102 and CIFAR-100 datasets, and mildly decreases after the optimal performances. The ensemble accuracies of the ERI scheme ($N=N_C$) on these two datasets are slightly lower than that of $N$-ary ECOC. Obviously, the ECOC fails to address the datasets with a large value of $N_C$, while the higher ensemble performance of $N$-ary ECOC needs a large value of $N$, namely, a small class merge degree. For the FLOWER-102 and CIFAR-100 datasets, the $N$-ary obtains good results after $N\geq 75$ and achieve best at $N=95$. In particular, the class merge degree for FLOWER-102 is $0.069$ and CIFAR-100 is $0.05$.

In general, we conclude from the experiment that the ensemble performance is relatively stable for the datasets with a small value of $N_C$, which slightly improves until the peak and then decreases a bit, or just slightly decreases with the increase of $N$ after the peak. While the performance, for the datasets with a large value of $N_C$, boosts significantly at the very beginning, then it saturates as $N$ continues increasing and reaching the optimum when $N$ is close to $N_C$. This could be explained by that the base learners with large $N$ has stronger discriminability~\cite{zhou2019n}.

Thus, our suggestions for the choice of $N$ are: 1) For the dataset with small $N_C$, the large class merge degree strategy, \ie small $N$, is better for achieving good performance, such as $N=3$ or $4$ for the dataset with $N_C\leq 10$. 2) Reversely, for the dataset with large $N_C$, the small class merge degree strategy should be applied, \eg $75\leq N\leq 95$ for $N_C$ is around $100$.

\begin{table}[t]
    \scriptsize
    \caption{\small Ensemble accuracies with their standard deviations.}
	\label{tab:esemble_acc_wrt_n_l}
	\setlength{\tabcolsep}{3.8 pt}
	\centering
	\begin{threeparttable}
	\begin{tabular}{l l c c c c c c c c}
		\toprule
        \multirow{2}{*}{Dataset} & \multirow{2}{*}{$N$} & \multicolumn{8}{c}{\# of Base Learners ($N_L$)} \\
        & & 10 & 20 & 30 & 45 & 50 & 60 & 80 & 100 \\
        \midrule
        MNIST & 3 & 99.14\% & 99.20\% & 99.35\% & 99.48\% & \textbf{99.57\%} & \textbf{99.57\%} & - & - \\
        CIFAR-10 & 4 & 87.45\% & 89.76\% & 91.78\% & 91.83\% & 91.82\% & 91.92\% & \textbf{91.95\%} & 91.93\% \\
        CIFAR-100 & 95 & 67.94\% & 69.12\% & 69.11\% & 69.33\% & 69.34\% & 69.46\% & 69.67\% & \textbf{69.94\%} \\
        FLOWER-102 & 95 & 86.06\% & 86.45\% & 86.45\% & 87.06\% & 87.16\% & \textbf{87.94\%} & 87.46\% & 87.59\% \\
        \midrule
        TREC & 3 & 93.80\% & 94.00\% & 95.20\% & 95.20\% & \textbf{95.60\%} & \textbf{95.60\%} & 95.50\% & \textbf{95.60\%} \\
        SST & 4 & 46.74\% & 48.19\% & 49.41\% & 50.18\% & 50.45\% & \textbf{50.86\%} & - & - \\
        \bottomrule
	\end{tabular}
	\begin{tablenotes}
        \small
        \item[] \scriptsize{*Here $N_L$ are $60$, $100$, $100$, $60$, $60$ and $60$, respectively, for the ensemble models from top to bottom row. While $N$ are $3$, $4$, $95$, $95$, $3$, $4$, respectively, for the $N$-ary ECOC.}
      \end{tablenotes}
     \end{threeparttable}
\end{table}

\subsubsection{Evaluation on the Effect of Base Learner Number $N_L$.}
\begin{figure}[t]
    \centering
	\includegraphics[width=0.9\textwidth]{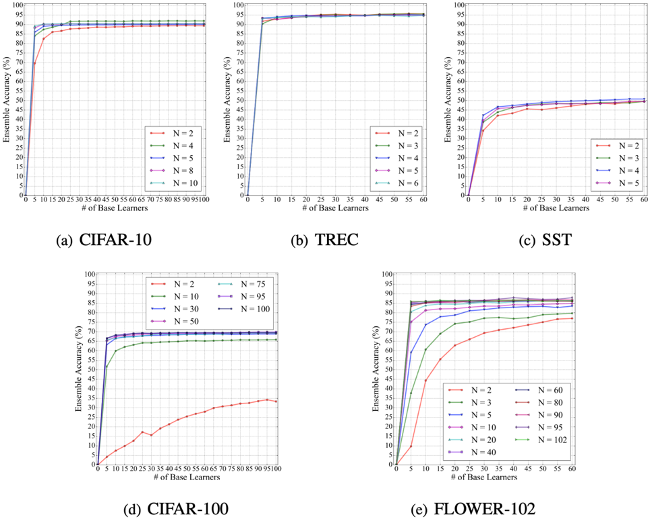}
	\caption{\small Ensemble accuracies of different values for $N$ with respect to $N_L$ on the image and text datasets, where $N=2$ is ECOC scheme, $N=N_C$ (biggest $N$ in each sub-figure) is ERI scheme, and the rest is $N$-ary ECOC schemes.}
	\label{fig:ensemble_acc_diff_n_wrt_n_l}
\end{figure}

\begin{figure}[t]
    \centering
	\includegraphics[width=0.9\textwidth]{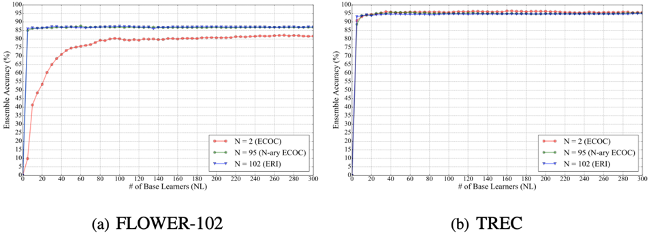}
	\caption{\small Ensemble accuracies with respect to large $N_L$($=300$) for three coding schemes on FLOWER-102 and TREC datasets.}
	\label{fig:ensemble_acc_wrt_large_n_l}
\end{figure}

In this experiment, we further explore another crucial hyper-parameter of $N$-ary ECOC, namely the number of base learner $N_L$ (also equivalent to the code length), and study its influence on the ensemble accuracy. We first report the ensemble accuracies of different $N_L$ for each dataset with the optimal meta-class number $N$, as described in Table~\ref{tab:esemble_acc_wrt_n_l}. Then, we study the ensemble accuracies of different meta-class $N$ with respect to $N_L$ (see Fig.~\ref{fig:ensemble_acc_diff_n_wrt_n_l} and~\ref{fig:ensemble_acc_wrt_large_n_l}).

From Table~\ref{tab:esemble_acc_wrt_n_l}, we observe that one requires a smaller number of base learners $N_L$ for datasets with small $N_C$ than that for datasets with large $N_C$ to reach the optimal ensemble accuracies generally. For example, MNIST and TREC only need $50$ base learners to get the optimum, while SST obtains best accuracies with $60$ base learners and CIFAR-10 requires $80$. In comparison, it reaches the optimal ensemble accuracies with the help of $100$ base learners on CIFAR-100 (note that it first reaches optimum when $N_L=90$). There is a special issue that FLOWER-102 holds a large $N_C$ ($102$ classes), but only requires $60$ base learners to derive the optimal ensemble accuracies. It is because the pre-trained model on the large-scale dataset (ILSVRC dataset in our experiment) is utilized and the pre-trained model already encodes a variety of abstractly and typically well-learned features. Moreover, we also find that the requirement of $N_L$ is related to the single model performance to some degree, say, the single model achieves better performance, then its ensemble model requires fewer base learners to achieve the optimal result.

In addition to the observations from Table~\ref{tab:esemble_acc_wrt_n_l}, we also study the impact of $N_L$ on ensemble performance under different meta-class $N$. Obviously, the optimal number of base learners $N_L$ for achieving the best accuracy is related to the meta-class $N$, as shown in Fig.~\ref{fig:ensemble_acc_diff_n_wrt_n_l} and~\ref{fig:ensemble_acc_wrt_large_n_l}. Normally, an ensemble model with small meta-class $N$ needs more base learners $N_L$ to achieve the same result compared to an ensemble model with large meta-class. It is because the discriminative ability of the codes for small $N$ is worse than that for large $N$.

Considering the definition of ECOC, $N$-ary ECOC and ERI schemes, the discriminative ability of ECOC is worst due to its small meta-class ($N=2$), which means ECOC needs relatively most base learners to reach the optimal performance compared with $N$-ary ECOC and ERI, where ERI holds the best discriminative ability. Thus, we conclude that $N_L$ for ECOC is greater than or equal to $N_L$ for $N$-ary ECOC, while $N_L$ for $N$-ary ECOC is greater than or equal to $N_L$ for ERI. Note that we use ``greater than or equal to'' since there is no guarantee that the optimal $N_L$ for a small $N$ must be larger than that for a large $N$, especially for some extreme situations such as $N=2$ (ECOC) versus $N=3$ ($N$-ary ECOC) or $N=99$ ($N$-ary ECOC) versus $N=100$ (ERI) for the dataset with $100$ classes ($N_C$).

In Fig.~\ref{fig:ensemble_acc_diff_n_wrt_n_l}, the experiment results on all the datasets show similar trends that ensemble accuracies of larger $N$ converge faster than that of smaller $N$ as the increasing of $N_L$, which means larger $N$ requires less $N_L$ and vice versa. For example, as shown in Fig.~\ref{fig:ensemble_acc_diff_n_wrt_n_l}(a), ECOC reaches optimal ensemble accuracies at $N_L=100$, while $N$-ary ECOC with $N=4$ and $5$ optima at $N_L=80$, then $N_L=60$ for $N=8$ and ERI peaks at $N_L=55$. The patterns on TREC and SST datasets are consistent as CIFAR-10. Typically, such patterns are more distinct for datasets with large $N_C$ (ref. Fig.~\ref{fig:ensemble_acc_diff_n_wrt_n_l}(d) and~\ref{fig:ensemble_acc_diff_n_wrt_n_l}(e)). For instance, in Fig.~\ref{fig:ensemble_acc_diff_n_wrt_n_l}(d), the ensemble accuracies of $N=10$ are highest at $N_L=100$, $N_L=95$ for $N$-ary ECOC with $N=30,50,75$, while $90$ base learners are required for $N=95$ and ERI needs $N_L=85$. Here we do not take ECOC into consideration, since it fails to improve the ensemble accuracy with only $34.26\%$. For the Fig.~\ref{fig:ensemble_acc_diff_n_wrt_n_l}(e), we see that ensemble accuracies of $N=2,3,5,10$ converge at $N_L=60$, $N=20,40$, converges at around $N_L=50$, $N=60,80,90$ at approximately $N_L=45$ and $40$ base learners are needed for $N$-ary ECOC with $N=95$ and ERI to reach the optimal ensemble accuracy.

Apart from the optimal $N_L$ for each meta-class $N$ to reach optimal ensemble accuracy, we also observe that using $15\sim 25$ base learners for ERI is good enough for datasets with small $N_C$ while $20\sim 40$ base learners for large $N_C$. For ECOC, it fails with the large $N_C$, and on the dataset with small $N_C$. Although ECOC performs comparably to $N$-ary ECOC and ERI, it still needs more base learners to converge, which is different from the conclusion in~\cite{allwein2000reducing} that ECOC requires $N_L=10\log_{2}(N_C)$ on traditional classifiers. For $N$-ary ECOC, the optimal performance is highly related to the choice of $N$. If the choice of $N$ follows suggestions in Section~\ref{sssec:meta_class}, $40\sim 60$ base learners for small $N_C$ are enough to achieve good performance, while large $N_C$ needs around $60\sim 100$ base learners.

We further extend number of base learners to $300$ and experiment on the FLOWER-102 and TREC datasets to investigate ensemble performances with the increasing $N_L$, as in Fig.~\ref{fig:ensemble_acc_wrt_large_n_l}. From Fig.~\ref{fig:ensemble_acc_wrt_large_n_l}(a), we find that the performance of ECOC improves significantly when $N_L$ increases, then keep relatively stable with a slight increase after $N_L=100$ and reach the optimal accuracy of $82.17\%$ at around $N_L=270$. However, the best performance of ECOC derived by using a large number of base learners is still lower than $N$-ary ECOC with $N=95$ and ERI with only $5$ base learners used, which indicates that ECOC is not suitable for the large $N_C$ case. For the $N$-ary ECOC and ERI, they obtain good scores with only small numbers of base learners and slightly improve to the optimal accuracy at around $N_L=40$. After that, the performance remains stable with the increase of $N_L$ and it drops when $N_L$ continues to increase, which indicates that increasing $N_L$ monotonously has no impact on performance. Similar observations could be found in Fig.~\ref{fig:ensemble_acc_wrt_large_n_l}(b).

Generally, there is no concrete conclusion for the choice of the number of base learners $N_L$, but some helpful guidelines can be summarized for experiments: 1) The choice of meta-class $N$ is more important than the number of base learners $N_L$ for the performance of $N$-ary ECOC, especially for the dataset with large $N_C$. Since the increase of $N_L$ cannot compensate for the negative effects caused by a badly selected $N$ (\eg $N=10$ for CIFAR-100). 2) Albeit the optimal number of base learners $N_L$ varies along $N_C$, the suggested $N_L$ is in the range of $\big[\lfloor 10\log_{2.2}(N_C)\rfloor, \lceil 10\log_{1.5}(N_C)\rceil\big]$. For example, the optimal $N_L$ ranges in $[30,58]$ for $N_C=10$ and $[59,110]$ for $N_C=100$, which aligns with the observations in our experiments.

\subsubsection{Comparison with Three Parameter Sharing Strategies.}
In this Section, we study the effect of three different parameter sharing strategies in the framework of ECOC, $N$-ary ECOC, and ERI. Note that, for the $N$-ary ECOC framework, we only select the optimal meta-class $N$ of each dataset for display except for the CIFAR-100 dataset which four different $N$ are chosen for display. We first study the performance of three different parameter sharing strategies on each tested dataset.

\begin{figure}[t]
    \centering
	\includegraphics[width=0.9\textwidth]{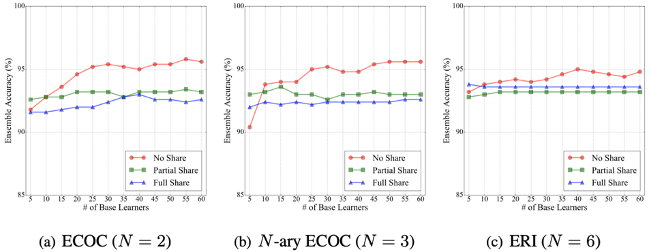}
	\caption{\small Parameter sharing strategies in ECOC, $N$-ary ECOC and ERI for TREC dataset.}
	\label{fig:param_sharing_trec}
\end{figure}

From the experimental results on the TREC dataset (see Fig.~\ref{fig:param_sharing_trec}), we observe that no parameter sharing strategy performs better than partial and full parameter sharing strategy for ECOC, N-ary ECOC, and ERI. When the number of base learner $N_L$ is small, the performance of no share is not satisfactory. Then it improves significantly with the increase of $N_L$, while the performances of partial and full share are relatively stable with respect to $N_L$. Moreover, when the number of meta-class $N$ is small, partial share outperforms the full share and the performance of no share is much better than partial and full share. However, when $N$ is large, full share is better than partial share and the performance of no share is just slightly higher than partial and full share.

\begin{figure}[t]
    \centering
	\includegraphics[width=0.9\textwidth]{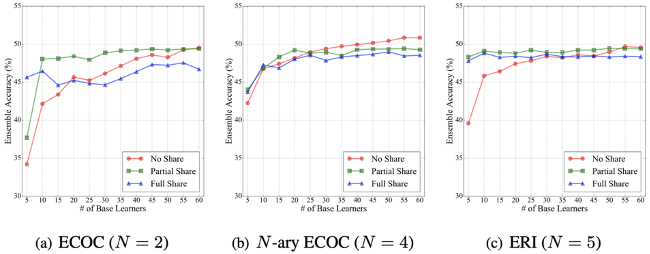}
	\caption{\small Parameter sharing strategies in ECOC, $N$-ary ECOC and ERI for SST dataset.}
	\label{fig:param_sharing_sst}
\end{figure}

From Fig.~\ref{fig:param_sharing_sst}, we have the following observations. First, when the number of meta-class $N$ is small, both partial and no share models improve significantly with the increase of $N_L$. The partial share generally outperforms the no and full share except when $N_L$ is less. Second, when the number of meta-class $N$ is large, as shown in Fig.~\ref{fig:param_sharing_sst}(b) and~\ref{fig:param_sharing_sst}(c), the performance of the three strategies are stable, and the improvement of no share is most significant with the increase of $N_L$. No share strategy governs the best performance with $N=4$ while partial share strategy always performs best for ERI situation.

\begin{figure}[t]
    \centering
	\includegraphics[width=0.9\textwidth]{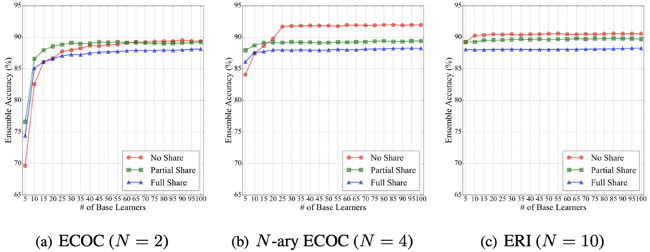}
	\caption{\small Parameter sharing strategies in ECOC, $N$-ary ECOC and ERI for CIFAR-10 dataset.}
	\label{fig:param_sharing_cifar10}
\end{figure}

\begin{figure}[t]
    \centering
	\includegraphics[width=0.9\textwidth]{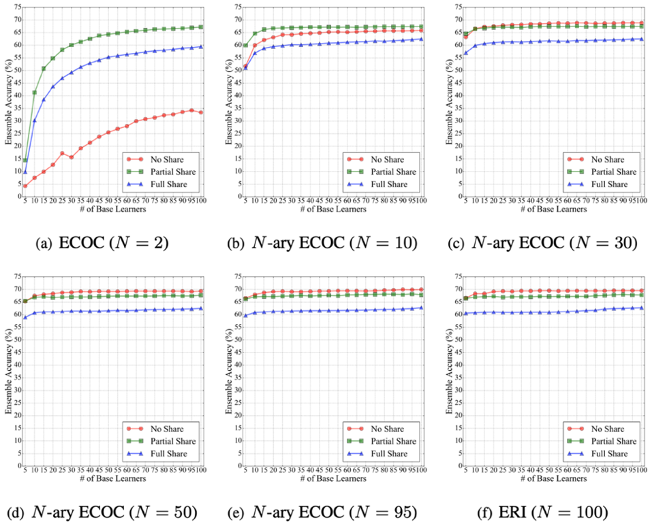}
	\caption{\small Parameter sharing strategies in ECOC, $N$-ary ECOC and ERI for CIFAR-100 dataset.}
	\label{fig:param_sharing_cifar100}
\end{figure}

In Fig.~\ref{fig:param_sharing_cifar10}, the performances of no, partial, and full share strategies are more stable. When the number of base learners $N_L$ is small, we see that the performance of no share is worst with ECOC and $N$-ary ECOC, and partial share performs better with $N$-ary ECOC and ERI situations. With the increase of $N_L$, for ECOC, all the strategies improve significantly, partial share outperforms another two strategies at the beginning, and then no share comes closer to partial share and reaches slightly higher performance than partial share. For $N$-ary ECOC, partial and full share strategies do not show significant improvement, while no share improves obviously and outperforms the partially and full share despite its lower ensemble accuracy at the very beginning. For the ERI, all these three strategies perform stable while no share always performs best and the performance of full share stays the bottom.

In the last experiment, we study the parameter sharing strategies in ECOC, $N$-ary ECOC, and ERI for the dataset with a large number of classes, as shown in Fig.~\ref{fig:param_sharing_cifar100}. For $N$-ary ECOC situation, we experiment on four different meta-class with $N=10,30,50,95$.

First, we observe that ECOC model with no share strategy fails to achieve satisfactory performance, while partial and full share strategies with the ECOC improve significantly with the increase of $N_L$. Moreover, partial share always outperforms full share.

Secondly, for the $N$-ary ECOC with small number of meta-class, we observe that partial share strategy outperforms no and full share always. No share improves most significantly and its performance is comparable to that of partial share with the increase of $N_L$. The performance of full share always maintains the worst. With an increasing number of meta-class $N$, partial share strategy outperforms no share strategy at the beginning, but its performance is gradually surpassed by no share when number of base learners $N_L$ increases. For $N=50$ and $95$, the performance of no share is comparable to that of partial share when the number of base learners $N_L$ is small. No share outperforms partial share with the increases of $N_L$. Moreover, for the $N$-ary ECOC, full share strategy consistently performs worst.

Thirdly, for the ERI model, the observations are similar to the $N$-ary ECOC with large meta-class $N$ and the no share strategy is comparable to partial share when $N_L$ is small. It always performs best when $N_L$ increases, meanwhile, the performance of full share is worst.

Finally, we conclude that: 1) In general, for the dataset with the small number of classes, the performance of no share model is better than or equal to that of the partial share model, thus no share strategy is suggested to be chosen. 2) For the dataset with the small number of classes, when the number of meta-class $N$ is large, these three strategies perform stable. 3) For the dataset with a large amount of classes, when the number of meta-class is small, the performance of partial share model is the best. 4) For the dataset with large amount of classes, when the number of meta-class is large, no share strategy model outperforms partial and full share models in most cases. Thus no share strategy should be preferred in such a case. 5) If the number of meta-class is large, the performance difference between three sharing strategies is marginal. Then full share could be suggested due to its parameter efficiency.

\section{Conclusion}\label{sec:conclude}
In this paper, we mainly investigate how to effectively integrate deep learning with the $N$-ary ECOC framework, also termed Deep $N$-ary ECOC. To achieve this goal, we give three different realizations. We further carry out extensive experiments to show the superiority of deep $N$-ary ECOC over existing data-independent deep ensemble strategies.

\section*{Acknowledgement}
The research work is supported by the Agency for Science, Technology and Research (A*STAR) under its AME Programmatic Funding Scheme (Project No. A18A1b0045). Ivor W. Tsang was supported by ARC DP180100106 and DP200101328.

%
%
\bibliographystyle{splncs04}
\bibliography{deep_nary_ecoc}
\end{document}